\documentclass[conference]{IEEEtran}
\IEEEoverridecommandlockouts
\usepackage{cite}
\usepackage{amssymb,amsfonts}
\usepackage{algorithmic}
\usepackage{textcomp}
\usepackage{xcolor}
\usepackage{cite}
\usepackage{booktabs} % For nicer tables
\usepackage{multirow} % For merged cells
\usepackage{array} % For better column definitions

\ifCLASSINFOpdf
  \usepackage[pdftex]{graphicx}
  \graphicspath{{../pdf/}{../jpeg/}}
  \DeclareGraphicsExtensions{.pdf,.jpeg,.png}
\else
  \usepackage[dvips]{graphicx}
  \graphicspath{{../eps/}}
  \DeclareGraphicsExtensions{.eps}
\fi

\usepackage{caption} % For customizing captions
\usepackage{subcaption} % For customizing captions

\usepackage[cmex10]{amsmath}

\usepackage{algorithmic}

\usepackage{array}

\usepackage{url}
\def\BibTeX{{\rm B\kern-.05em{\sc i\kern-.025em b}\kern-.08em
    T\kern-.1667em\lower.7ex\hbox{E}\kern-.125emX}}

\IEEEoverridecommandlockouts
\IEEEpubid{\makebox[\columnwidth]{979-8-3503-6491-0/24/\$31.00~\copyright2024 IEEE \hfill}
\hspace{\columnsep}\makebox[\columnwidth]{ }}

\begin{document}

\title{Transformer models as an efficient replacement for statistical test suites to evaluate the quality of random numbers}

\author{\IEEEauthorblockN{1\textsuperscript{st} Rishabh Goel}
\IEEEauthorblockA{\textit{Monta Vista High School} \\
Cupertino, United States \\
goelr668@gmail.com}
\and
\IEEEauthorblockN{2\textsuperscript{nd} Yizi Xiao}
\IEEEauthorblockA{\textit{Optum} \\
Eden Prairie, United States \\
joe.xiao@optum.com}
\and
\IEEEauthorblockN{3\textsuperscript{rd} Ramin Ramezani}
\IEEEauthorblockA{\textit{Department of Computer Science} \\
\textit{UCLA}\\
Los Angeles, United States \\
raminr@ucla.edu}
}

\maketitle

\IEEEpubidadjcol

\begin{abstract}
  Random numbers are incredibly important in a variety of fields, and the need for their validation remains important for safety. A Quantum Random Number Generator (QRNG) can theoretically generate truly random numbers, however their quality still needs to be thoroughly validated. Generally, the task of validating random numbers has been delegated to different statistical tests such as the tests from the NIST Statistical Test Suite (STS), which are often slow and only perform one test at a time. Our work presents a deep learning model utilizing the Transformer architecture that 1) performs multiple NIST STS tests at once, and 2) runs much faster. This model outputs multi-label classification results on passing these statistical tests. We performed a thorough hyper-parameter optimization to converge on the best possible model and as a result, achieved a high degree of accuracy with a Macro F1-score of above 0.96. We also compared this model to a conventional deep learning method (Long Short Term Memory Recurrent Neural Networks) to quantify randomness and showed our model achieved similar performances while being much more efficient and scalable. The high performance and efficiency of this Transformer-based deep learning model showed that it can be a viable replacement for the NIST STS for validating random numbers.  
\end{abstract}
\noindent\hspace{2.0em}\textbf{Index Terms}---Transformers, Deep learning, Random numbers, Multi-label classification, LSTMs

% Introduction
\section{Introduction}
Random numbers serve an important purpose in many fields with numerous applications. Within cryptography, they are extensively studied and utilized to ensure secure encryption schemes that keep our data safe \cite{corrigan2013ensuring,shen2020randomness,gennaro2006randomness,dodis2004possibility}. In physics, random numbers are highly studied in their appearances in quantum mechanics and thermodynamics \cite{natal2021entropy, bera2017randomness}. Given the vast array of applications associated with random numbers, their validation also becomes a critical issue\cite{li2020deep}. Most of our encryption schemes today use pseudo-random numbers generated through pseudo-random number generators, and their validation has been delegated to a variety of statistical tests \cite{arman2020design}.

Along with using statistical tests for validating random numbers, deep learning and its derivatives can be used for determining the randomness of random numbers \cite{9187201}. The uses of deep learning in this area have branched out to many of its facets such as Convolutional Neural Networks (CNNs), Recurrent Neural Networks (RNNs), and Long Short Term Memory RNNs (LSTMs) \cite{nagy2021randomness, sokorac2017optimizing, hochreiter1997long}. 

More recently, the introduction of the Transformer model has allowed further exploration of this deep learning method in the context of validating random numbers\cite{vaswani2017attention}. In the applications of the Transformer models, the self-attention head is designed to detect sequential patterns\cite{xu2019self}. Thus, in many applications where sequential data is available and easy to represent, the Transformer has been quite successful, especially in the field of Natural Language Processing (NLP) but also in many others \cite{khan2022transformers,lin2022survey,patwardhan2023transformers,geneva2022transformers,bran2023transformers}. 

Since a random number by definition would have minimal sequential patterns and a non-random number would be the opposite, we reasoned that the self-attention mechanism in the Transformer architecture should be quite effective at quantifying the randomness of random numbers which was confirmed by Li et al \cite{10405286}. In the context of random numbers, traditionally, LSTMs were a popular deep learning method for qualifying the randomness of random numbers, but due to their complex nature as well as their inability to be parallelized, they were slow and quite inefficient but were quite accurate \cite{zeyer2019comparison}. As Li et al. described, these problems with LSTMs were solved by the Transformer architecture. Although Li et al. showed that the Transformer architecture is well equipped for prediction of the next token and use that as a metric of validating random numbers, we focus on the quantification on the quality of randomness more directly through the encoding of various statistical tests rather than predicting what the next random number will be. 

We hypothesized that a deep learning architecture based on Transformers would be more efficient and high in performance at quantifying the randomness of random numbers as well as the type of randomness to a degree in which it could be considered as a potential replacement for the NIST STS. In this paper, we present an encoder-only Transformer model that can accurately encode and replace the statistical tests that were used to quantify the quality of random numbers. The goal is to develop a single model that transforms binary sequences into probabilities of passing different statistical tests, thereby quantifying the amount of randomness of the sequence as well as the type of randomness. We experimented with different architectures of this model as well as different hyperparameters and converged on the most optimized architecture for this problem. We showed that by utilizing the Transformer architecture, our model can accurately describe the type and degree of randomness of a binary sequence with comparable performance to an LSTM while improving efficiency over the original NIST STS and LSTM, serving as a strong potential replacement. 

\section{Materials and Methods}

Non-deep learning based methods of validating random numbers include using a variety of statistical test suites such as the DIEHARD and DIEHARDER statistical test suites along with the NIST STS \cite{brown2018dieharder,marsaglia2008marsaglia,rukhin2001statistical}. These suites contain a multitude of tests, each testing for a particular type of randomness. The purpose of this work is to encode a portion of these tests into a single Transformer in order to test random numbers on multiple different tests at the same time. 

\subsection{Dataset}
The training data for our model was first generated using truly random numbers from a Quantum Random Number Generator (QRNG), then augmented to have non random numbers as well as random numbers in the dataset. We ran the NIST STS to generate labels for our data. Running the NIST STS on each binary sequence would generate a p-value for each sequence. We used the alpha values as outlined by Rukhin et al. in the definition of the NIST STS as thresholds for determining if the sequence would pass each particular test. Running seven tests on each sequence would result in a one-hot encoded label vector for each input sequence. The purpose of the model is to run multi-label classification on given data so that the user can find out which tests the random number passes and which ones it fails. Thus, our model outputs a vector where each entry would be the probability of the random number passing the corresponding test. There were seven tests from the NIST STS that we encoded: Frequency, Block Frequency, Runs, Longest Run of Ones, Discrete Fourier Transform, Nonperiodic Template Matchings, Cumulative Sums. Out of the seven, we augmented data to generate non-random results for five of the tests: Frequency, Block Frequency, Runs, Longest Run of Ones, Nonperiodic Template Matching. We found that augmenting the data only for these five aforementioned tests also resulted in the binary sequences failing the Discrete Fourier Transform and Cumulative Sums tests, which we did not augment the data for. Therefore, it was not necessary to augment the data to introduce non-randomness for these tests in order for the model to encode them. The specific details of how these tests work are outlined clearly by Ruhkin et al. in their definition of NIST STS\cite{rukhin2001statistical}. The data augmentation techniques were adopted from Nagy et al. with minimal changes\cite{nagy2021randomness}.

\subsection{Model Training and Validation}
\begin{figure}[ht]
    \centering
    \includegraphics[width=0.45\textwidth]{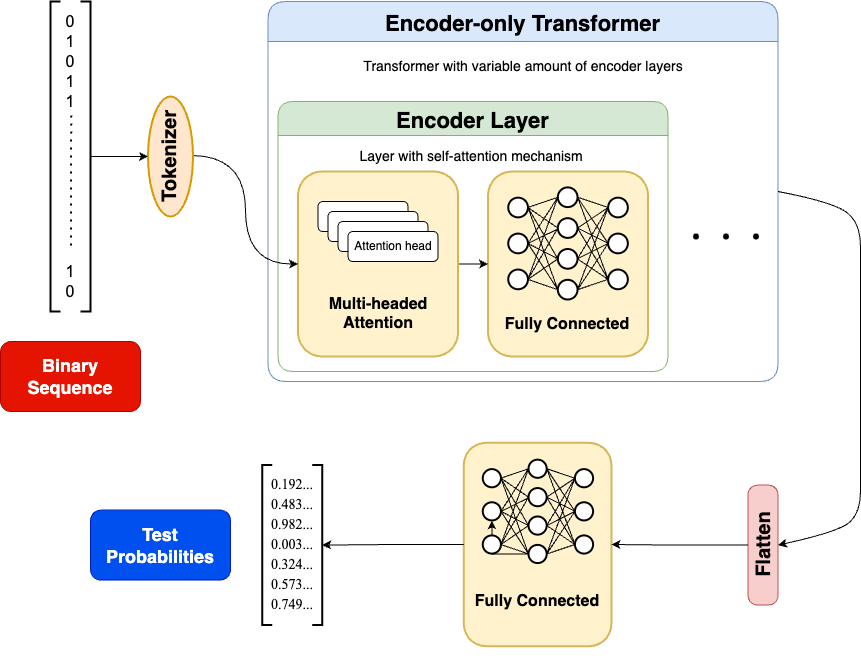}
    \caption{Baseline model architecture. The tokenizer includes positional encoding and embeddings. The flatten layer is swapped with the averaging layer when constructing the final model.}
    \label{fig:architecture}
\end{figure}

We began by taking the binary sequence and tokenizing it to reduce its dimensionality. We trained many models to experiment with hyper-parameters which included varying the input size. We used inputs of size 512, 1024, and 2048 bits to train our models due to their relevance in cryptography \cite{alese2012comparative} as well as resource constraints barring us from using input sizes past 2048 bits. Our tokenization technique was simply getting the integer representation of every 16 bits. This would create a vocabulary for the Transformer of size 65536 which was sufficiently large for model training. This means that our input would reduce to 32, 64, or 128 tokens, which we would then run through the model whose architecture is shown above in Figure \ref{fig:architecture}. 

Model training was done with three datasets, each with a size of 100000 labeled binary sequences pertaining to each input size with a 60-20-20 train-validation-test split. To monitor how well the model was training, we used aggregate F1 to measure model performance per batch as well as loss that best pertains to multi-label classification \cite{benedict2021sigmoidf1,yacouby2020probabilistic}. We use a combination of sigmoid, binary cross entropy loss with the Adam optimizer with default parameters as defined by Kingma et al. \cite{kingma2014adam} to train the model, and Macro, Micro, Weighted, and Sample F1 scores to validate it with equations available in the Appendix \ref{equations}. All training, testing, and analysis was performed on a single computer with the CPU: i5-9600k and GPU: 3060 ti. 
 
\subsection{Handling Varying Input Size} \label{sec:averaging}
\begin{figure}[ht]
    \centering
    \includegraphics[width=0.489\textwidth]{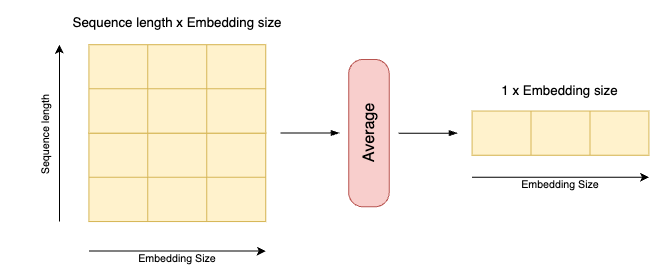}
    \caption{The effects on the shape of the input are illustrated. The averaging layer averages along each column to reduce the sequence length dimension to 1. The result is a vector that is the length of the embedding size. This vector is then passed into the fully connected layer as in Figure 1.}
    \label{fig:averaging}
\end{figure}

One of the great advantages of Transformer models is their ability to handle data with no fixed input size in parallel \cite{zhang2022vsa}. However, as seen in Figure \ref{fig:architecture}, since the output of the encoder is then flattened and connected into a fully connected layer, the input size must be fixed as a fully connected layer of an ANN cannot handle varying input size. 
Our solution to this was quite simple. Since the fully connected layer can only take a fixed input length, we collapsed the only dimension that was varying which is the sequence length dimension. The drawback of this of course is that there is some information lost through the averaging process and how this affects model performance is discussed later.

\subsection{Experimental Procedure}
There are three main hyper-parameters that we observed the performance of to find the best architecture: the number of encoder layers in the transformer, the number of heads in the multi-headed attention, the dimensionality or size of the embeddings, the input size, and whether or not we used an averaging layer. To get the data points for our analysis and to converge on the best possible model through hyper-parameter optimization, we used the following procedure:

\begin{enumerate}
    \item For each encoder layer number/embedding size/attention head number:
    \begin{enumerate}
        \item Train a model for each input size (512, 1024, 2048) and averaging type (averaging or non-averaging) with the corresponding architecture with a default of 3 encoder layers, 8 attention heads, and an embedding size of 240. (If you are testing for the number of attention heads, the number of encoder layers the model will have for all numbers will be 3 and the embedding size will be 240 for all numbers)
        \item Record all aggregate metrics.
    \end{enumerate}
    \item Determine the optimal architecture, input size, and averaging type.
    \item Train the model with the previously found optimal architecture and compare against LSTM for validation, recording all the aggregate metrics including time.
\end{enumerate}
% Results

\section{Results}
Our goal in this study is to create the best possible model that could be used as an alternative to the NIST STS in classifications of random numbers. To do this, we had to first converge on the best possible model utilizing the transformer architecture and then validate it using previously accepted techniques, namely LSTMs. The concrete values of the hyper-parameter optimization are too many to include in tabular form here so they may be found in supplemental form in the Appendix \ref{appendix}. We visualize and discuss the conclusions of the raw data in the Discussion and Analysis portion of this article. We found the optimal model for the task of classifying random numbers to be: one encoder layer, single-headed attention, 192 embedding size with the averaging layer. 

The next step was to validate this model with previously accepted techniques of classifying random numbers which are LSTMs and of course, the STS itself. Below are the tables providing the raw metrics of LSTM and Transformer inference performance and time as well as comparisons to the STS.

\begin{table}[h]
    \centering
    \caption{Performance metrics for input size 512}
    \label{tab:512}
    \begin{tabular}{lccccc}
        \toprule
        \textbf{Technique} & \textbf{Inference Time (s)} & \textbf{Micro F1} & \textbf{Macro F1} \\ 
        \midrule
        LSTM         & 3.046 & \textbf{0.931} & 0.932 \\
        Transformer  & \textbf{0.965} & \textbf{0.931} & \textbf{0.934} \\
        STS          & 3.82 & - & -\\
        \bottomrule
    \end{tabular}
\end{table}

% Table for Input Size 1024
\begin{table}[h]
    \centering
    \caption{Performance metrics for input size 1024}
    \label{tab:1024}
    \begin{tabular}{lccccc}
        \toprule
        \textbf{Technique} & \textbf{Inference Time (s)} & \textbf{Micro F1} & \textbf{Macro F1} \\ 
        \midrule
        LSTM         & 5.172 & \textbf{0.967} & \textbf{0.970} \\
        Transformer  & \textbf{1.071} & 0.961 & 0.962 \\
        STS          & 4.63 & - & - \\
        \bottomrule
    \end{tabular}
\end{table}

% Table for Input Size 2048
\begin{table}[h]
    \centering
    \caption{Performance metrics for input size 2048}
    \label{tab:2048}
    \begin{tabular}{lccccc}
        \toprule
        \textbf{Technique} & \textbf{Inference Time (s)} & \textbf{Micro F1} & \textbf{Macro F1} \\ 
        \midrule
        LSTM         & 8.991 & \textbf{0.965} & \textbf{0.964} \\
        Transformer  & \textbf{1.201} & 0.960 & \textbf{0.964} \\
        STS          & 5.73 & - & - \\
        \bottomrule
    \end{tabular}
\end{table}

\section{Discussion and Analysis}

We set out with the goal of creating the best possible model that can be used as an alternative to the statistical test suites. To find the model with the best hyper-parameters, we ran tests with different hyper-parameter settings to converge on the best possible model that we can find with our current architecture. The hyper-parameters that we tested for were the number of encoder layers, the length of the embedding dimension, and the number of attention heads in the multi-headed attention section of the encoder. For brevity's sake, we are only showing the Macro F1 scores and omitting the other aggregate measures. To validate our model, we also compared it to previous methods of classifying the type of randomness and quantifying the degree of randomness, namely NIST's STS and LSTMs. Here we take the raw data presented above and build visualizations that contextualize the data in our study.

\subsection{Comparative Analysis}

% \begin{figure}[h]
%     \centering
%     \begin{subfigure}[b]{0.475\textwidth}
%         \includegraphics[width=\textwidth]{comparison.png}
%         \label{fig:performance}
%     \end{subfigure}
%     \hfill 
%     \begin{subfigure}[b]{0.425\textwidth}
%         \includegraphics[width=\textwidth]{time.png}
%         \label{fig:time}
%     \end{subfigure}
%     \caption{LSTM vs Encoder-only Transformer model. On the left: Macro F1 performance comparison of our Transformer based model compared to the more widely used LSTM architecture. On the right: Inference time of the Transformer model versus the NIST STS.  Each model and the STS had to run through a test set of 20000 binary sequences for each input size.}
%     \label{fig:comparisons}
% \end{figure}
\begin{figure}[ht]
    \centering
    \includegraphics[width=0.489\textwidth]{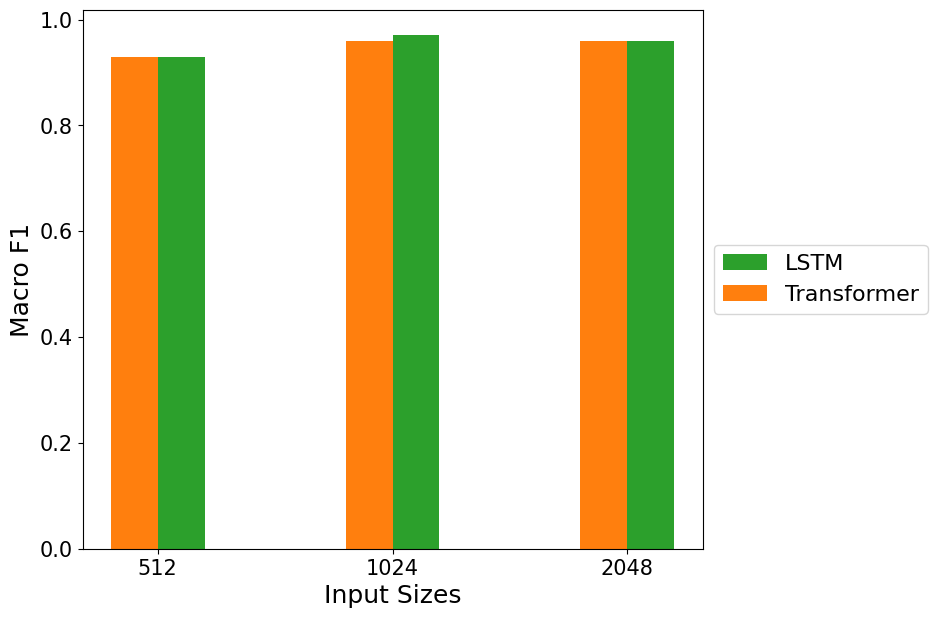}
    \caption{Performance of our Transformer based model compared to the more widely used LSTM architecture.}
    \label{fig:performance}
\end{figure}

As outlined by Nagy et al. and Li et al. traditional RNN based methods for classifying random numbers have been using LSTMs. To compare Transformer model performance with LSTM performance, we used the LSTM architecture as outlined by Nagy et al. with the addition of embeddings and positional encoding as that improved performance. As seen above, the Transformer based approach achieves similar levels of performance as compared to the LSTM with the LSTM being marginally better at times. Of course, this benefit is not justified by the time needed to both train and run the model which is discussed later. Overall, their performances are similar enough to be deemed comparable and interchangeable. Furthermore, it does not appear as if input size made a difference at all in the performance of either the LSTM or the Transformer. This can have two main explanations. First, it is likely that with much larger input sizes (perhaps an order of magnitude greater), the pitfall of LSTMs being poor handlers of long input sizes might become more evident. This however requires more exploration and such lengthy input sizes were not tested due to our testing machine being unable to handle these lengthy inputs. Second, perhaps the task of detecting randomness and quantifying it in binary sequences does not require long term sequential dependencies that Transformers are better suited than LSTMs to handle.

\begin{figure}[ht]
    \centering
    \includegraphics[width=0.45\textwidth]{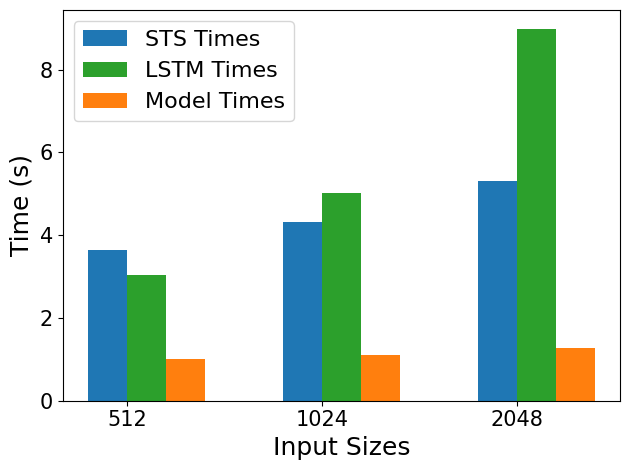}
    \caption{Runtime of the Transformer model versus the NIST STS. Each model and the STS had to run through a test set of 20000 binary sequences for each input size.}
    \label{fig:time}
\end{figure}

Discussed further on, we have performed a hyper-parameter optimization and even explored a range of values of each hyper-parameter to find trends in model performance and we have converged on a model with 1 encoder layer with single-headed attention with an embedding dimension of 192. To further explore the usability of this model, we compared the time taken to run through our test dataset for both the NIST STS as well as our model. Since our model can run on the GPU and the Transformer architecture is parallelizable, it runs much faster than the NIST STS which runs on the CPU \cite{pope2023efficiently}. Our model processes the same amount of data as the STS in almost 33\% of the time. As can be seen, the LSTM is considerably slower due to it not being as parallelizable as the Transformer model. As input size grows, the LSTM begins to take considerably more time, even slower than STS, while the Transformer seems to grow at a much slower rate and never even comes close to the STS time. The usability of LSTM drops dramatically as input size increases and clearly LSTM should not even be considered as a replacement for STS given the lack of any observable efficiency gain. 

\subsection{Encoder Layers}

\begin{figure}[h]
    \centering
    \begin{subfigure}[b]{0.24\textwidth}
        \includegraphics[width=\textwidth]{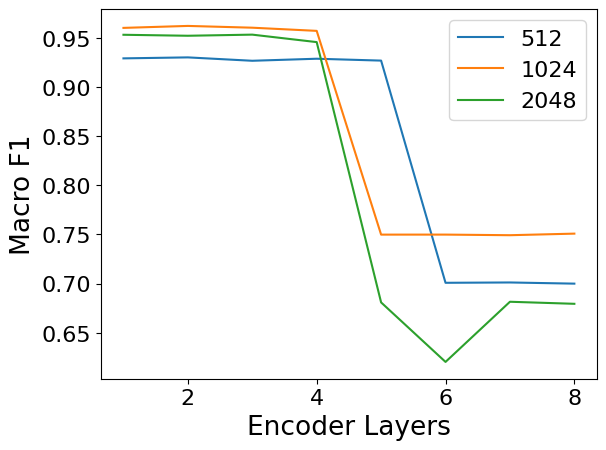}
        \caption{With averaging layer}
        \label{fig:image1}
    \end{subfigure}
    \hfill
    \begin{subfigure}[b]{0.24\textwidth}
        \includegraphics[width=\textwidth]{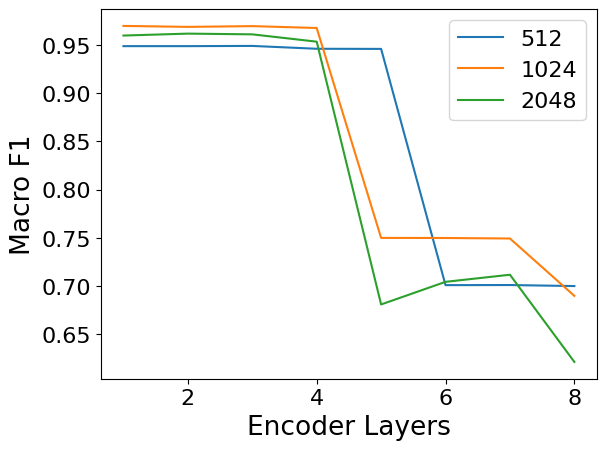}
        \caption{Without averaging layer}
        \label{fig:image2}
    \end{subfigure}
    \caption{Encoder Layers versus Macro F1 scores}
    \label{fig:encoders}
\end{figure}

As Huang et al. show, more Transformer encoder layers should improve the model accuracy and in our case, F1 score \cite{huang2020improving}. While we did observe this to be true, the benefits were marginal and there were some important caveats. Namely, the effects were only beneficial until a certain point; in our case, once the number of encoder layers exceeded three, the model stopped learning and failed to converge as seen in Figure \ref{fig:encoders}. A possible conclusion that can be drawn from this is that the model became too large for the input and ended up adding too much noise for any useful information to be perceptible in later layers. However, it may also be the case that the amount of data we had was not sufficient to train a model with more than four encoder layers, regardless of whether we were averaging across the outputs or not. From this graph however, we can say that one encoder layer is sufficient for this task and any more encoder layers would only cause an increase in training time and inference time without providing any tangible performance benefit.

\subsection{Embedding Size}

\begin{figure}[h]
    \centering
    \begin{subfigure}[b]{0.24\textwidth}
        \includegraphics[width=1\textwidth]{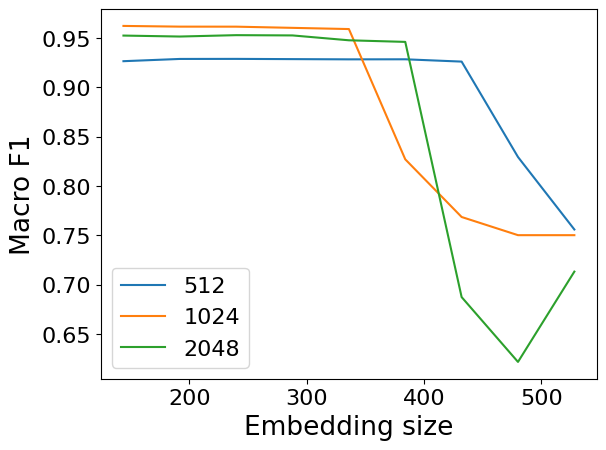}
        \caption{With averaging layer}
        \label{fig:img3}
    \end{subfigure}
    \hfill
    \begin{subfigure}[b]{0.24\textwidth}
        \includegraphics[width=1\textwidth]{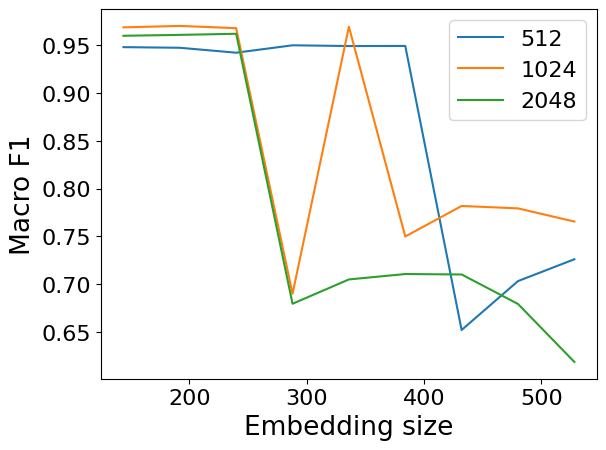}
        \caption{Without averaging layer}
        \label{fig:img4}
    \end{subfigure}
    
    \caption{Embedding dimension size versus Macro F1 scores}
    \label{fig:inputs}
\end{figure}

The length of the embedding of the tokenized input string is crucial to the performance of the model but again we can see that there is not much of a difference until the embedding size gets around 375 (250 for the non-averaging version)\cite{huang2020improve}. At this point, the embedding size becomes too high and no model is able to converge to an acceptable F1 score (above 0.7). Furthermore, it seems like input size plays a marginal role in performance with the model performing slightly better with longer input sizes (1024 and 2048 bit sequences). However, what is interesting is that the 512 bit input seems to be more resilient so to say, as it converges for a greater range of embedding sizes compared to the other input sizes. Again, without the averaging layer, the F1 score ceiling is higher with the 1024 bit stream reaching F1 scores of above 0.95 however this difference again seems to be only marginal as the top F1 scores of the models with the averaging layers also exceeding 0.95 however by a smaller amount. The main observations are that the averaging layer allows the models to converge for a larger embedding size and that the smaller the input size, the larger the embedding size it converges for. Once more, there is not a tangible performance increase as the embedding size increases so an embedding size of 192 (the smallest we tested) is sufficient for the task of classifying random numbers. Since the non-averaging models and the models with higher input sizes seem to be performing worse, it is likely that they require more data in order to train but more exploration on this matter is needed. 

\subsection{Number of Attention Heads}

\begin{figure}[ht]
    \centering
    \begin{subfigure}[b]{0.24\textwidth}
        \includegraphics[width=1\textwidth]{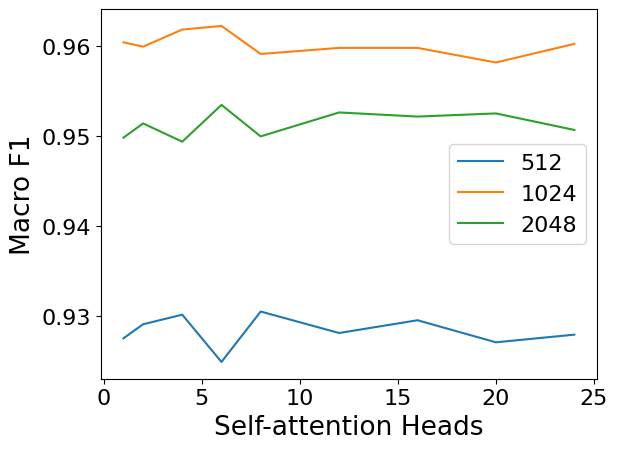}
        \caption{With averaging layer}
        \label{fig:img5}
    \end{subfigure}
    \hfill
    \begin{subfigure}[b]{0.24\textwidth}
        \includegraphics[width=1\textwidth]{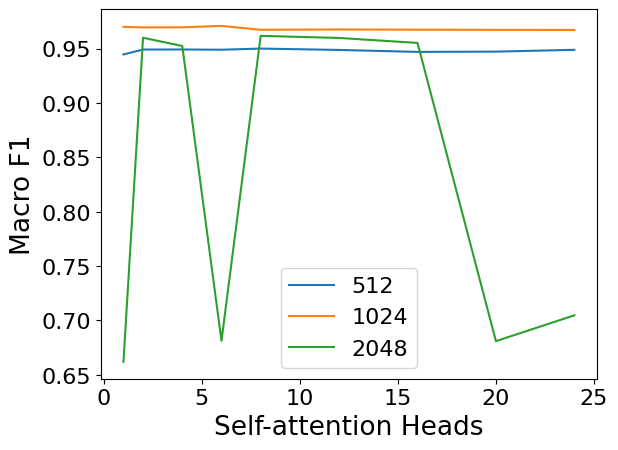}
        \caption{Without averaging layer}
        \label{fig:img6}
    \end{subfigure}
    
    \caption{Number of attention heads versus Macro F1 scores}
    \label{fig:heads}
\end{figure}

The number of attention heads in the multi-headed attention layer is important to finding the most optimized model as the number of attention heads can quite readily affect the performance of a model \cite{merity2019single}. Therefore, we also experimented with the number of multi-headed attention layers to find the most optimized model. As seen in Figure \ref{fig:heads}, the number of heads did not play a large role in the performance of the model at all with more attention heads only being marginally better but adding quite a lot of overhead. As a common trend, we see that the averaging layer did lower overall Macro F1, yet it was more stable as the 2048 input size model failed to converge past 20 self-attention heads layers. Relatively speaking, input sizes of 32, 64 and 128 tokens are quite small and therefore do not require large models to compute which is why more self-attention heads likely did not positively impact performance. However what was unexpected was that without the averaging layer, the 2048 input size model failed to converge with larger self-attention heads. Perhaps at this stage a greater restructuring of the model architecture would be required or a larger dataset. Through this graph, the conclusion can be drawn that single headed attention is sufficient for a model that does average across the output of the encoder and that two attention heads are sufficient for a non-averaging model. As before, there is no real tangible performance gain so more attention heads would only lead to longer training and inference times. Overall, it seems that the averaging layer provides stability and resilience to over-training, making it preferable to the non-averaging layer despite it being marginally worse in terms of performance. This performance deficit is made up, however, by the fact that the averaging layer allows the model to handle varying input sizes, thus allowing it to be trained on all three datasets at once. Combined with its added resilience to over-training, the performance of the model with the averaging layer actually ended up being better than without. 

\section{Conclusion}

In conclusion, we see that deep learning models that utilize the Transformer architecture are adequate alternatives for the tests of the NIST STS and serve as a faster and scalable alternative as our highest performing model can classify over 20000 numbers in almost 50\% of the time it takes the NIST STS to do so (perhaps even faster with a more powerful GPU). With the use of the averaging layer, our model can handle inputs of varying size from 512 bits to 2048 bits, showing versatility in its capabilities. Our work also invites further exploration by expanding the model's capabilities to encoding the rest of the statistical tests presented in NIST STS, further expanding the applicability of our model. Overall, our model serves as a promising potential replacement for the NIST STS given its time efficiency and comparable performance to more traditional deep learning methods such as LSTMs.

\bibliographystyle{IEEEtran}
\newpage
\bibliography{references.bib}

\appendix
\subsection{Concrete data values}
\label{appendix}

All concrete data values used in the visualizations can be found on the Github: https://github.com/bloodpool7/RandomTransformer
% replace [link] with [\href{https://github.com/bloodpool7/RandomTransformer}{link}]

\subsection{F1 Scores}
\label{equations}
Here are the equations used to define the f1 metrics which were used to validate our model.

\noindent
\\
\textbf{Micro F1 Score}:
\begin{equation}
    \frac{{2 \times \text{Micro Precision} \times \text{Micro Recall}}}{{\text{Micro Precision} + \text{Micro Recall}}}
\end{equation}

\noindent
\\
\textbf{Macro F1 Score}:
\begin{equation}
    \frac{{1}}{{N}} \sum_{i=1}^{N} \frac{{2 \times \text{Precision}_i \times \text{Recall}_i}}{{\text{Precision}_i + \text{Recall}_i}}
\end{equation}

\noindent
\\
\textbf{Weighted F1 Score}:
\begin{equation}
    \frac{{1}}{{\text{Total Support}}} \sum_{i=1}^{N} (\text{Support}_i \times \text{F1}_i)
\end{equation}

\noindent
\\
\textbf{Sample F1 Score}:
\begin{equation}
    \frac{{2 \times \text{True Positives}}}{{2 \times \text{True Positives} + \text{False Positives} + \text{False Negatives}}}
\end{equation}

\end{document}